\documentclass[sigconf]{acmart}
\usepackage{graphicx}
\usepackage{subfig}
\usepackage{geometry}
\usepackage{multirow}

\AtBeginDocument{%
  }

\copyrightyear{2025}
\acmYear{2025}
\setcopyright{cc}
\setcctype{by}
\acmConference[KDD '25]{Proceedings of the 31st ACM SIGKDD Conference on Knowledge Discovery and Data Mining V.2}{August 3--7, 2025}{Toronto, ON, Canada}
\acmBooktitle{Proceedings of the 31st ACM SIGKDD Conference on Knowledge Discovery and Data Mining V.2 (KDD '25), August 3--7, 2025, Toronto, ON, Canada}
\acmDOI{10.1145/3711896.3737439}
\acmISBN{979-8-4007-1454-2/2025/08}

\begin{document}

\title{\texttt{TimeGraph}: Synthetic Benchmark Datasets for Robust Time-Series Causal Discovery}

\author{Muhammad Hasan Ferdous}
\affiliation{
\institution{Causal AI Lab}
  \institution{University of Maryland, Baltimore County}
  \city{Baltimore}
  \state{Maryland}
  \country{USA}
}
\email{h.ferdous@umbc.edu}

\author{Emam Hossain}
\affiliation{
\institution{Causal AI Lab}
  \institution{University of Maryland, Baltimore County}
  \city{Baltimore}
  \state{Maryland}
  \country{USA}
}
\email{emamh1@umbc.edu}

\author{Md Osman Gani}
\affiliation{
\institution{Causal AI Lab}
  \institution{University of Maryland, Baltimore County}
  \city{Baltimore}
  \state{Maryland}
  \country{USA}
}
\email{mogani@umbc.edu}

\renewcommand{\shortauthors}{Ferdous et al.}

\begin{abstract}
Robust causal discovery in time series datasets depends on reliable benchmark datasets with known ground-truth causal relationships. However, such datasets remain scarce, and existing synthetic alternatives often overlook critical temporal properties inherent in real-world data, including nonstationarity driven by trends and seasonality, irregular sampling intervals, and the presence of unobserved confounders. To address these challenges, we introduce \texttt{TimeGraph}, a comprehensive suite of synthetic time-series benchmark datasets that systematically incorporates both linear and nonlinear dependencies while modeling key temporal characteristics such as trends, seasonal effects, and heterogeneous noise patterns. Each dataset is accompanied by a fully specified causal graph featuring varying densities and diverse noise distributions and is provided in two versions: one including unobserved confounders and one without, thereby offering extensive coverage of real-world complexity while preserving methodological neutrality. We further demonstrate the utility of TimeGraph through systematic evaluations of state-of-the-art causal discovery algorithms including PCMCI+, LPCMCI, and FGES across a diverse array of configurations and metrics. Our experiments reveal significant variations in algorithmic performance under realistic temporal conditions, underscoring the need for robust synthetic benchmarks in the fair and transparent assessment of causal discovery methods. The complete TimeGraph suite, including dataset generation scripts, evaluation metrics, and recommended experimental protocols, is freely available to facilitate reproducible research and foster community-driven advancements in time-series causal discovery. Source code is available at \url{https://github.com/hferdous/TimeGraph}.

\end{abstract}

\begin{CCSXML}
<ccs2012>
   <concept>
       <concept_id>10002950.10003648.10003688.10003693</concept_id>
       <concept_desc>Mathematics of computing~Time series analysis</concept_desc>
       <concept_significance>500</concept_significance>
       </concept>
   <concept>
       <concept_id>10002950.10003648.10003649.10003655</concept_id>
       <concept_desc>Mathematics of computing~Causal networks</concept_desc>
       <concept_significance>500</concept_significance>
       </concept>
   <concept>
       <concept_id>10002944.10011123.10011130</concept_id>
       <concept_desc>General and reference~Evaluation</concept_desc>
       <concept_significance>500</concept_significance>
       </concept>
   <concept>
       <concept_id>10002944.10011123.10011124</concept_id>
       <concept_desc>General and reference~Metrics</concept_desc>
       <concept_significance>500</concept_significance>
       </concept>
 </ccs2012>
\end{CCSXML}

\ccsdesc[500]{Mathematics of computing~Time series analysis}
\ccsdesc[500]{Mathematics of computing~Causal networks}
\ccsdesc[500]{General and reference~Evaluation}
\ccsdesc[500]{General and reference~Metrics}

\keywords{causal discovery benchmarks, synthetic time series, temporal causality, ground truth temporal datasets, multivariate time series, non-stationary data, missing data}

\maketitle

\section{Introduction}

Causality forms the foundation for many scientific pursuits, providing a framework for understanding how changes in one component of a system can affect outcomes elsewhere \cite{pearl2009causality}. Such insights are important for robust decision-making, strategic interventions, and reliable predictions in fields as diverse as economics, healthcare, and engineering. In contrast to mere association or correlation, causal analysis seeks to identify structural relationships that govern a system's behavior, offering explanatory power that extends well beyond predictive accuracy. 

Within this domain, \textit{causal discovery} focuses on the identification of causal structures—often represented as directed graphs—that encode the directional relationships and causal dependencies among variables from observational data \cite{spirtes2000causation}. Correctly identifying these causal structures is paramount. For instance, in economics, understanding the causal impact of policy changes on economic indicators can guide effective governance. In healthcare, discerning the true causal pathways between lifestyle factors, treatments, and patient outcomes is crucial for developing effective interventions and personalized medicine. Misidentifying a correlational relationship as causal can lead to flawed policies, ineffective treatments, and a fundamental misunderstanding of the system under study.

\begin{figure}[htbp]
    \centering
    \includegraphics[width=0.37\textwidth]{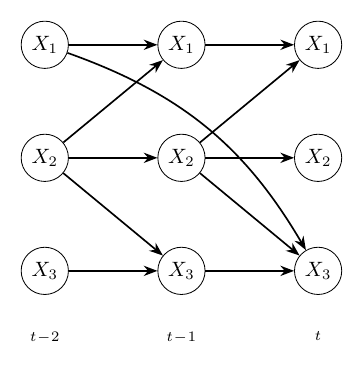}
    \Description{A time-series causal graph with variables shown at time points t, t-1, and t-2. Arrows illustrate both autoregressive and cross-lagged causal dependencies.}
    \caption{Illustrative time-series causal graph. Nodes denote variables at times $t$, $t\!-\!1$, and $t\!-\!2$. Arrows show autoregressive and cross-lagged causal directions.}
    \label{fig:simple_ts_causal_graph}
\end{figure}

The challenges and importance of causal discovery are amplified when analyzing \textit{time-series data}. Temporal data introduces complexities such as lagged effects (where a cause at time $t-\tau$ influences an effect at time $t$), feedback loops (where variables influence each other over time), and dynamic interactions that evolve \cite{granger1969investigating}. For example, in climate science, understanding the lagged causal influence of greenhouse gas emissions on global temperature changes is critical for accurate modeling and prediction. Figure~\ref{fig:simple_ts_causal_graph} illustrates a simple time-series causal graph, where past values of one variable can influence the current or future values of itself or other variables. Accurately identifying these temporal causal links allows for more precise forecasting, better control strategies in engineering systems, and a deeper understanding of dynamic processes across various domains.

Although a variety of theoretically grounded algorithms exist, empirical progress is often obstructed by the absence of benchmark datasets containing verifiable ground-truth causal structures. This shortcoming hampers fair evaluations, fosters inconsistent reporting, and complicates comparisons across different approaches. This challenge is particularly acute because, while generating time-series data is relatively straightforward, constructing these datasets with verifiable, ground-truth causal structures especially those integrating multiple realistic temporal complexities remains a significant hurdle for robust algorithm development and evaluation.

The challenges become more pronounced when analyzing \emph{time-series} data, which introduce temporal complexities like lags, feedback loops, and seasonal or trending patterns \cite{granger1969investigating}. Real-world time-series rarely disclose the \emph{true} underlying causal mechanisms, and their complexity can either obscure or mimic causality. Synthetic datasets with established causal relationships can fill this gap by providing controlled experimental environments \citep{pearl2009causality,spirtes2000causation}. However, many existing synthetic benchmarks are designed with assumptions that may not reflect real-world complexities \citep{granger1969investigating}. In some cases, researchers tailor synthetic data to highlight the strengths of their proposed approaches, which can yield more favorable performance on these benchmarks \citep{mooij2016distinguishing}. Typically, these datasets rely on linear or stationary assumptions, for example, basic vector autoregressive models with Gaussian noise \citep{shimizu2006linear}, thereby overlooking the nonstationary or nonlinear phenomena prevalent in actual processes. Many papers attempt to introduce nonlinearity using polynomial transformations, such as squaring or cubing variables \citep{zhang2012kernel}, but in real-world time series nonlinearity often emerges from trends or seasonal fluctuations \citep{granger1969investigating}. 
Consequently, while such synthetic datasets may demonstrate promising results in controlled settings, they risk failing when applied to real-world cases or different datasets, ultimately limiting their utility as comprehensive evaluations of causal discovery methods \citep{runge2019}.

To tackle these limitations, we introduce \textbf{TimeGraph}, a synthetic time-series benchmark suite specifically designed to support rigorous and fair evaluations of causal discovery algorithms. TimeGraph models realistic temporal behaviors including trends, seasonality, regime changes, and heterogeneous noise while supporting complex causal patterns. TimeGraph's primary contribution lies not just in incorporating these individual features, but in its systematic integration of these complexities within a unified framework, offering known causal ground truths for diverse scenarios, a resource currently lacking in the field. In addition, we provide open-source generation scripts and standardized evaluation protocols to promote reproducibility and transparent cross-method comparisons. In this paper, we make the following key contributions:

\begin{itemize}
    \item We propose a principled methodology for generating synthetic time-series data that incorporate realistic temporal features (e.g., trends, seasonality) and multiple cross-lag dependencies among variables, thereby reflecting the complexities of real-world scenarios more faithfully than purely linear or stationary designs.

    \item We demonstrate the unique value of TimeGraph by providing a benchmark where complex temporal dynamics coexist with explicitly defined and controllable causal ground truths, including variants with latent confounders, enabling a more rigorous and fair assessment than benchmarks derived from real data where the true structure is unknown.

    \item We conduct extensive empirical evaluations of multiple state-of-the-art causal discovery algorithms under these realistic conditions, illustrating how trends, nonstationarity, nonlinearities, and potential confounders can affect performance.

    \item We release the \textbf{TimeGraph} benchmark suite—comprising generation scripts, detailed documentation, and standardized evaluation guidelines—to enable reproducible research and ensure consistent comparisons, thereby providing a more robust foundation for future developments in time-series causal discovery.
\end{itemize}

\noindent
The remainder of this paper is structured as follows:  
Section~\ref{sec:related} reviews time-series causal discovery benchmarks.  
Section~\ref{sec:variants} details our synthetic dataset variants and temporal behaviors.  
Section~\ref{sec:realism} examines dataset realism and benchmarking utility.  
Section~\ref{sec:benchmark} presents empirical results across algorithms under varying conditions.  
Section~\ref{sec:guidelines} offers best practices for future research.  
Section~\ref{sec:limitations} discusses limitations, and Section~\ref{sec:conclusion} outlines future directions.

\section{Related Work} \label{sec:related}

Causal discovery has evolved significantly since its early foundations in econometrics with Granger causality \cite{granger1969investigating} and the development of graphical models and do-calculus by Pearl \cite{pearl2009causality}. These seminal works established frameworks for inferring causal relationships from observational data. The classic approaches primarily focused on linear relationships and stationary processes, leading to methods like structural equation models and constraint-based algorithms \cite{spirtes2000causation}.

Time series causal discovery presents unique challenges beyond static data analysis. Early methods extended Granger causality to nonlinear settings \cite{chen2004analyzing} and developed specialized frameworks for temporal data \cite{peters2013causal}. Recent advances have addressed non-stationarity \cite{huang2020causal, ferdous2023cdans, ferdous2023ecdans}, heterogeneous sampling \cite{runge2019}, and complex temporal dependencies \cite{malinsky2019learning}. These techniques have been successfully applied to high-dimensional real-world systems, such as Arctic climate modeling, where integrating MVGC and PCMCI+ with deep learning improves the prediction and interpretability of sea ice dynamics \cite{hossain2024time, hossain2025correlation}. However, the systematic evaluation of these methods has been hindered by the lack of comprehensive benchmark datasets that incorporate realistic temporal characteristics \cite{HasanHG23, gong2024causal}.

Recognizing this scarcity, some researchers have attempted to establish ground-truth causal graphs from real-world observational data by leveraging extensive domain knowledge, established scientific theories, and expert opinions. For instance, in econometrics, structural models often impose restrictions based on economic theory to identify causal effects from time series \cite{blanchard1988dynamic}. Similarly, in fields like systems biology, causal models are sometimes constructed and validated against pathways derived from years of accumulated experimental evidence and expert consensus \cite{sachs2005causal}. Specific applications in clinical research also demonstrate the augmentation of data-driven causal models with expert knowledge, for example, to estimate treatment effects in intensive care units \cite{Gani2023}. Even in complex domains like climate science, evaluating discovered causal links often involves comparison against known physical mechanisms \cite{runge2020discovering}. While these approaches provide valuable, context-rich causal hypotheses rooted in reality, the resulting graphs often represent a consensus or a strong prior rather than an objectively verifiable ground truth suitable for rigorously testing the limits of diverse causal discovery algorithms. Furthermore, these expert-derived graphs may not always capture the full complexity or all latent interactions present in the system, and their availability is limited to specific, well-studied domains.

Several benchmark datasets have been proposed for causal discovery evaluation. The Cause-Effect Pairs database \cite{mooij2016distinguishing} provided a collection of real and simulated cause-effect relationships, while the CausalWorld environment \cite{ahmed2020causalworld} offered a physics-based simulation platform. More recently, CausalBench \cite{wang2024causalbench} established a standardized evaluation framework for causal learning algorithms. However, these resources primarily focus on static relationships or simplified temporal dynamics, often neglecting key characteristics of real-world time series data. CausalTime \cite{cheng2023causaltime} addresses the realism gap by starting from real observations, then using deep neural networks and normalizing flows to create more naturalistic time-series benchmarks, validated via PCA, t-SNE, discriminative metrics, and MMD. However, it relies on importance-based or domain-informed graph extraction and does not explicitly handle latent confounding, thus introducing potential biases and limiting applicability for certain causal discovery methods.

The treatment of missing data in causal discovery has gained increasing attention \cite{tu2019causal}, with researchers developing methods to handle both Missing Completely at Random (MCAR) and Missing Not at Random (MNAR) mechanisms \cite{gain2018structure}. However, existing benchmark datasets rarely incorporate realistic missing data patterns, particularly in the context of time series where missingness often exhibits temporal structure. Additionally, the presence of unmeasured confounders poses a fundamental challenge in causal discovery \cite{pearl2009causality}. While methods have been developed to detect and account for latent confounding \cite{zhang2018causal}, evaluation datasets typically either ignore this aspect or treat it in oversimplified ways.

Our TimeGraph framework addresses these limitations by providing a comprehensive suite of synthetic time series datasets that incorporate irregular sampling intervals, non-stationary dynamics, realistic trends and seasonal patterns, mixed noise distributions, structured missing data mechanisms (both MCAR and block-missing), and time-varying latent confounders. This enables systematic evaluation of causal discovery methods under conditions that more closely approximate real-world challenges.

\begin{figure}[t]
    \centering
    \includegraphics[trim=150 0 0 0, clip, width=\linewidth]{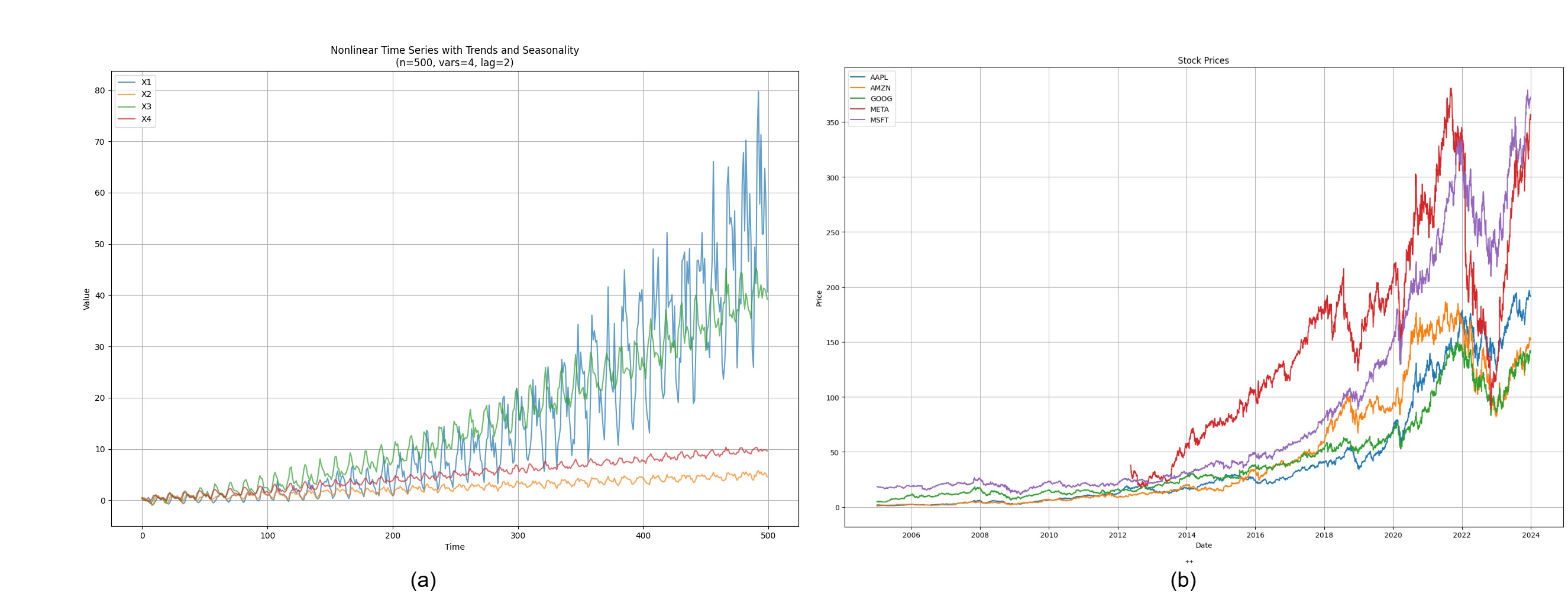}
    \Description{Side-by-side comparison of a synthetic time series and a real-world stock time series. Both show upward trends, cyclic behavior, and short-term fluctuations.}
    \caption{Comparison of our synthetic time series (C1 variant, left) with real-world stock data (right). Both exhibit qualitative similarities, such as strong upward trends, cyclical patterns, and short-term fluctuations, illustrating TimeGraph's ability to model key types of market-like temporal dynamics for benchmarking purposes. See Table~\ref{tab:synthetic-datasets} for C1 details.}
    \label{fig:c1_and_stock}
\end{figure}

\begin{table*}[htbp]
\centering
\small
\caption{Overview of the synthetic dataset variants. Each dataset is defined by its underlying functional form, error distribution, temporal sampling scheme, missing data pattern, and the presence of latent (unobserved) confounders. All datasets are generated with a fixed maximum lag (from \(\{2,3,4\}\) time steps) and at multiple sample sizes (500, 1,000, 3,000, and 5,000) to assess the robustness and scalability of causal discovery algorithms.}
\resizebox{\textwidth}{!}{%
\begin{tabular}{|c|p{2.0cm}|p{2.0cm}|p{2.0cm}|p{2.0cm}|p{1.7cm}|p{4.2cm}|}
\hline
\textbf{Dataset ID} & \textbf{Functional Form} & \textbf{Error Distribution} & \textbf{Temporal Sampling} & \textbf{Missing Data Pattern} & \textbf{Latent Confounders} & \textbf{Description/Notes} \\ \hline
\multicolumn{7}{|l|}{\textbf{Linear Time Series}} \\ \hline
A1   & Linear               & Gaussian and t         & Regular   & Complete     & No  & Baseline linear SEM with uniformly spaced observations. \\ \hline
A1C  & Linear               & Gaussian and t         & Regular   & Complete     & Yes & A1 with confounder \(U\) added (e.g., \(U\to X_3\) and \(U\to X_1\)). \\ \hline
A2   & Linear               & Gaussian and t          & Irregular & Complete     & No  & Linear SEM with heavy-tailed noise and non-uniform inter-arrival times. \\ \hline
A2C  & Linear               & Gaussian and t          & Irregular & Complete     & Yes & A2 with \(U\) incorporated to complicate causal inference. \\ \hline
\multicolumn{7}{|l|}{\textbf{Nonlinear Time Series (Polynomial)}} \\ \hline
B1   & Polynomial           & Gaussian and t                & Regular   & Complete     & No  & Nonlinear effects via quadratic and cubic terms with regular sampling. \\ \hline
B1C  & Polynomial           & Gaussian and t                & Regular   & Complete     & Yes & B1 augmented with \(U\) (quadratic terms) to induce hidden nonlinear dependencies. \\ \hline
B2   & Polynomial           & Mixed (Gaussian \& Laplace) & Irregular & Complete  & No  & B1 with mixed errors and irregular sampling. \\ \hline
B2C  & Polynomial           & Mixed (Gaussian \& Laplace) & Irregular & Complete  & Yes & B2 with \(U\) added to further obscure causal relationships. \\ \hline
\multicolumn{7}{|l|}{\textbf{Time Series with Trend \& Seasonality}} \\ \hline
C1   & Trigonometric with Trend \& Seasonality & Gaussian and t   & Regular   & Complete     & No  & Equations include sine/cosine trend-seasonal components (deterministic trends and \(T=12\) seasonal patterns). \\ \hline
C1C  & Trigonometric with Trend \& Seasonality & Gaussian and t   & Regular   & Complete     & Yes & C1 with latent confounders affecting selected variables via quadratic terms. \\ \hline
C2   & Trigonometric with Trend \& Seasonality & Gaussian and t    & Irregular & Complete     & No  & Same as C1 but with irregularly spaced timestamps. \\ \hline
C2C  & Trigonometric with Trend \& Seasonality & Gaussian and t   & Irregular & Complete     & Yes & C2 with an added confounder \(U\) to further challenge causal discovery. \\ \hline
\multicolumn{7}{|l|}{\textbf{Linear with MCAR Missingness}} \\ \hline
D1   & Linear (A1 variant)  & Gaussian and t               & Regular   & MCAR         & No  & A1 with random (MCAR) missing data simulating sensor dropouts. \\ \hline
D1C  & Linear (A1 variant)  & Gaussian and t               & Regular   & MCAR         & Yes & D1 with \(U\) included, adding hidden correlations to the missing data scenario. \\ \hline
\multicolumn{7}{|l|}{\textbf{Nonlinear with Block Missingness}} \\ \hline
D2   & Nonlinear (B1 variant) & Gaussian and t            & Irregular & Block Missingness (MAR/NMAR) & No  & B1 with contiguous blocks of missing data mimicking systematic outages. \\ \hline
D2C  & Nonlinear (B1 variant) & Gaussian and t            & Irregular & Block Missingness (MAR/NMAR) & Yes & D2 with \(U\) to assess causal discovery under block missingness and hidden confounding. \\ \hline
\multicolumn{7}{|l|}{\textbf{Nonlinear with Mixed Errors \& Mixed Missing Patterns}} \\ \hline
D3   & Trend \& Seasonality (C1 variant) & Mixed (Gaussian \& Laplace) & Irregular & Combined (MCAR + Block) & No  & Integrates trend/seasonality with mixed noise and dual missingness patterns. \\ \hline
D3C  & Trend \& Seasonality (C1 variant) & Mixed (Gaussian \& Laplace) & Irregular & Combined (MCAR + Block) & Yes & D3 with confounders, representing the most challenging scenario with multiple complexity sources. \\ \hline
\end{tabular}%
}
\label{tab:synthetic-datasets}
\end{table*}

\section{Overview of Synthetic Dataset Variants} \label{sec:variants}

In this study, we introduce \textbf{TimeGraph}, a comprehensive suite of synthetic time-series datasets designed to benchmark causal discovery algorithms systematically. Each dataset is generated with a known ground-truth causal structure and incorporates complexities observed in real-world time series, including nonlinearity, irregular sampling, latent confounding, and missing data mechanisms. By systematically varying these factors, TimeGraph allows researchers to evaluate the strengths and weaknesses of causal discovery methods under diverse conditions.

Each dataset variant is defined by its functional form, noise distribution, temporal sampling scheme, and missing data pattern. In particular, we construct two versions for each dataset family: one with fully observed variables and one with a latent confounder \( U \), which introduces hidden correlations among variables. These confounded variants simulate the challenges posed by unmeasured factors in real-world systems, making them especially useful for evaluating methods that claim robustness against hidden confounding. Additionally, all datasets are generated across multiple sample sizes (500, 1,000, 3,000, and 5,000) to analyze the effect of data volume on causal discovery performance. Table~\ref{tab:synthetic-datasets} summarizes the dataset variants and their defining characteristics.

\textbf{Linear Time Series (A1, A1C).}  
The A1 datasets represent the simplest case, where variables evolve according to a linear structural equation model (SEM) with fixed-lagged dependencies \citep{pearl2009causality, spirtes2000causation}. The base equations define directed, time-lagged relationships between variables, and errors are drawn from either a Gaussian or Student’s \( t \)-distribution to introduce variability. This dataset serves as a fundamental benchmark for causal discovery methods that assume linearity and additive noise.  

The confounded variant (A1C) extends this setup by introducing an unobserved confounder \( U \) that influences multiple variables. This simulates the common real-world scenario where unmeasured external factors introduce spurious correlations, making it difficult to distinguish direct from indirect causal effects. In practical applications, such latent confounding often arises in economic modeling, epidemiology, and finance, where external market forces or environmental conditions act as hidden common causes \citep{angrist2009mostly, hernan2010causal}. Failure to account for such confounders can lead to incorrect policy recommendations, misattribution of causal effects, and flawed risk assessments in predictive modeling.

\textbf{Nonlinear Time Series with Polynomial Functions (B1, B1C).}  
The B1 datasets extend the linear framework by incorporating polynomial transformations, introducing nonlinear dependencies among variables. Instead of simple linear interactions, variables now exhibit quadratic and cubic relationships, leading to complex functional structures that challenge traditional causal discovery algorithms \citep{mooij2016distinguishing, zhang2012identifiability, peters2017elements}. The non-monotonic transformations in B1 make it difficult for simple linear regression-based causal discovery methods to recover the true causal relationships, providing a critical benchmark for evaluating nonlinear causal discovery techniques.

In the confounded version (B1C), the latent variable \( U \) interacts nonlinearly with selected variables via quadratic terms. This represents cases where a hidden variable introduces a nonlinear bias into the system, such as in climate modeling, where an unobserved environmental factor (e.g., ocean temperature) affects multiple atmospheric variables in a nonlinear manner. Nonlinear confounding is also common in genetics, where complex epistatic interactions between genes can influence multiple traits in a non-additive manner. If these interactions are not explicitly accounted for, causal discovery methods may misattribute direct genetic effects to observed correlations, leading to incorrect conclusions about heritability and gene function.

\begin{figure}[t]
    \centering
    % First row
    \subfloat[Configuration 1: \(n=500\), vars=4, lag=2.]{%
        \includegraphics[width=0.3\linewidth]{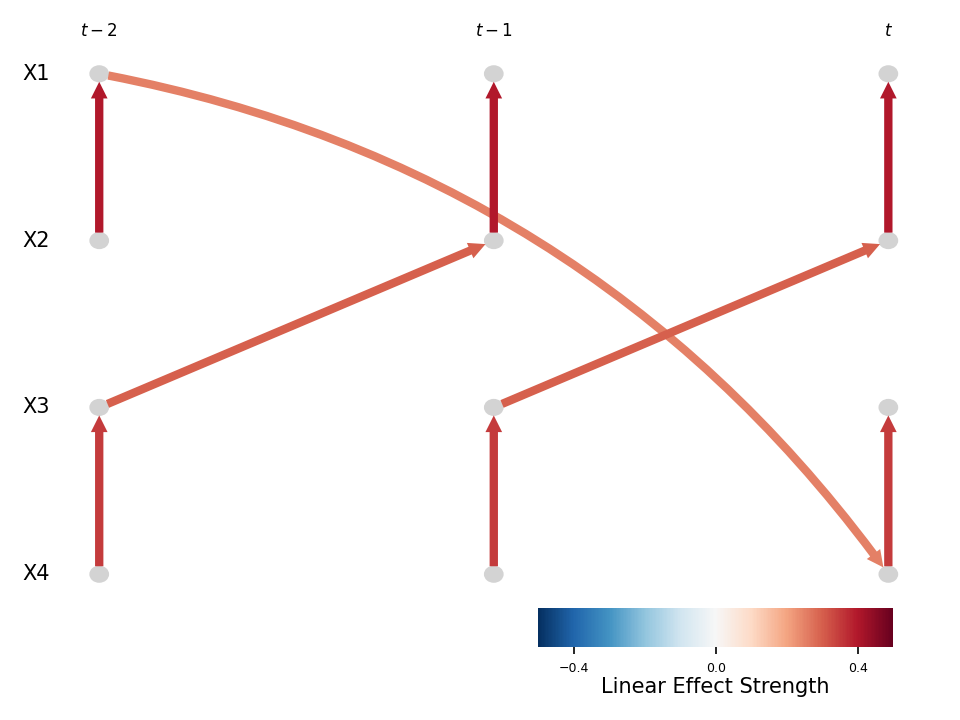}%
        \Description{Configuration 1: Causal graph with 4 variables, lag 2, 500 samples.}%
        \label{fig:cfg1}}
    \hfill
    \subfloat[Configuration 2: \(n=500\), vars=4, lag=3.]{%
        \includegraphics[width=0.3\linewidth]{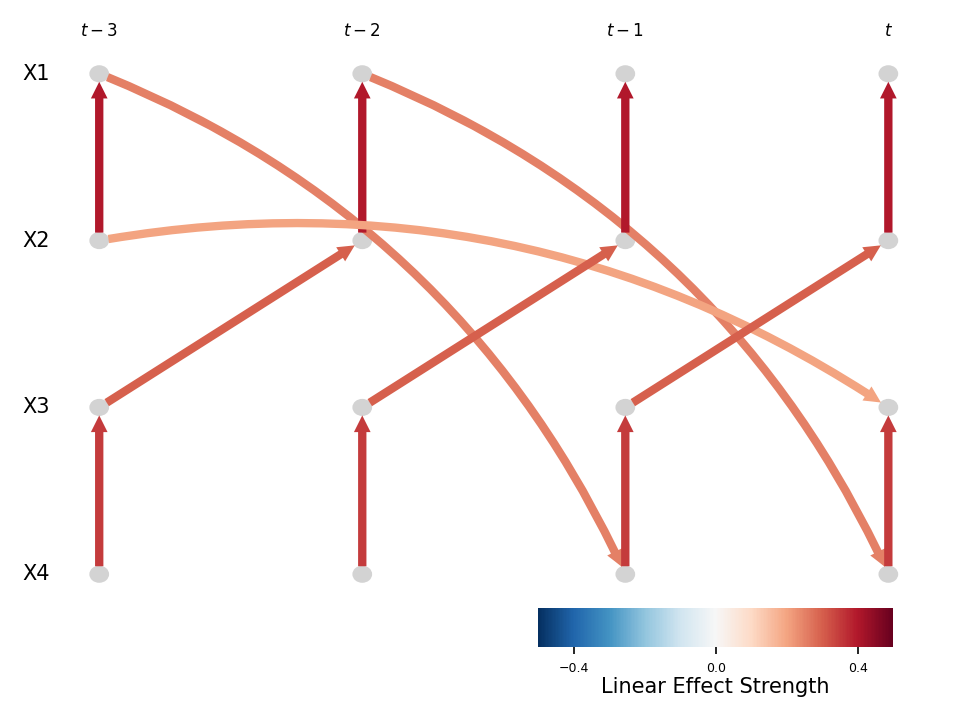}%
        \Description{Configuration 2: Causal graph with 4 variables, lag 3, 500 samples.}%
        \label{fig:cfg2}}
    \hfill
    \subfloat[Configuration 3: \(n=500\), vars=4, lag=4.]{%
        \includegraphics[width=0.3\linewidth]{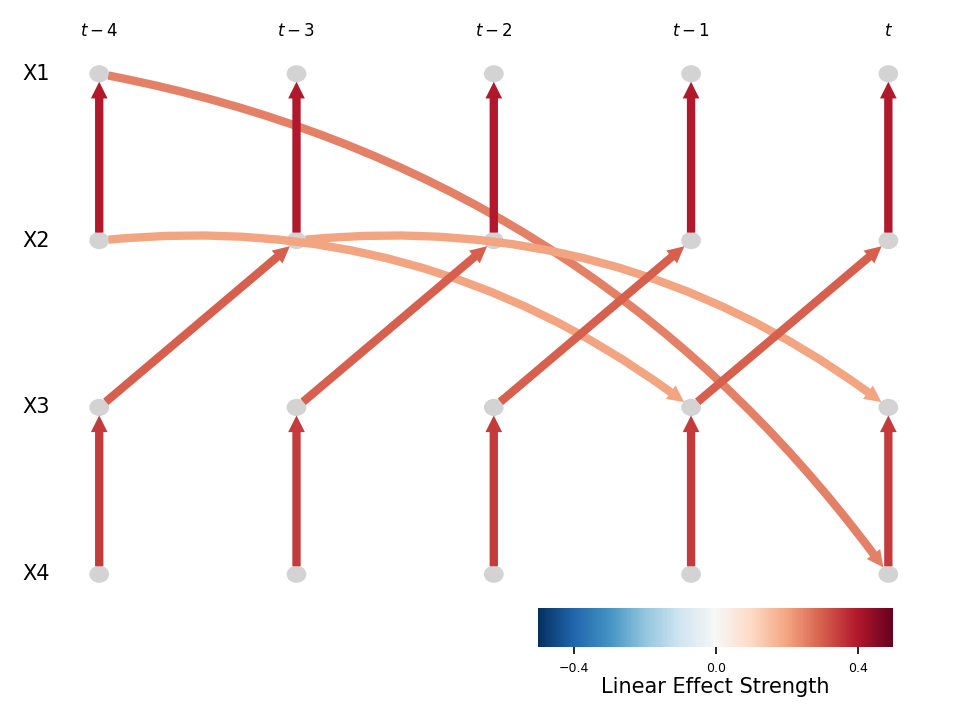}%
        \Description{Configuration 3: Causal graph with 4 variables, lag 4, 500 samples.}%
        \label{fig:cfg3}}

    \vspace{1em}

    % Second row
    \subfloat[Configuration 4: \(n=500\), vars=6, lag=2.]{%
        \includegraphics[width=0.3\linewidth]{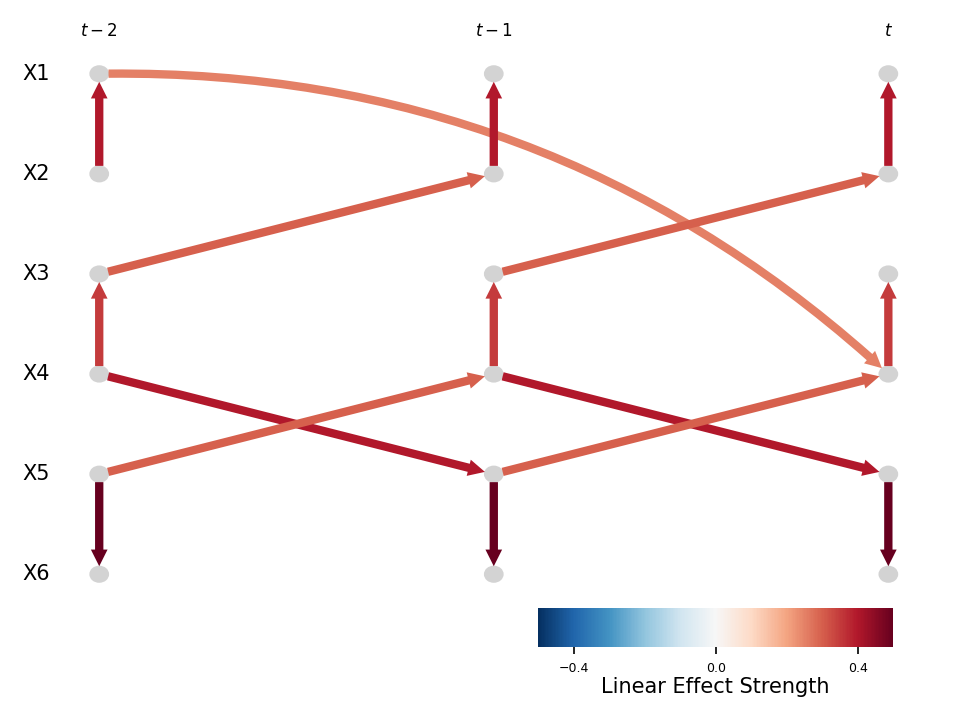}%
        \Description{Configuration 4: Causal graph with 6 variables, lag 2, 500 samples.}%
        \label{fig:cfg4}}
    \hfill
    \subfloat[Configuration 5: \(n=500\), vars=6, lag=3.]{%
        \includegraphics[width=0.3\linewidth]{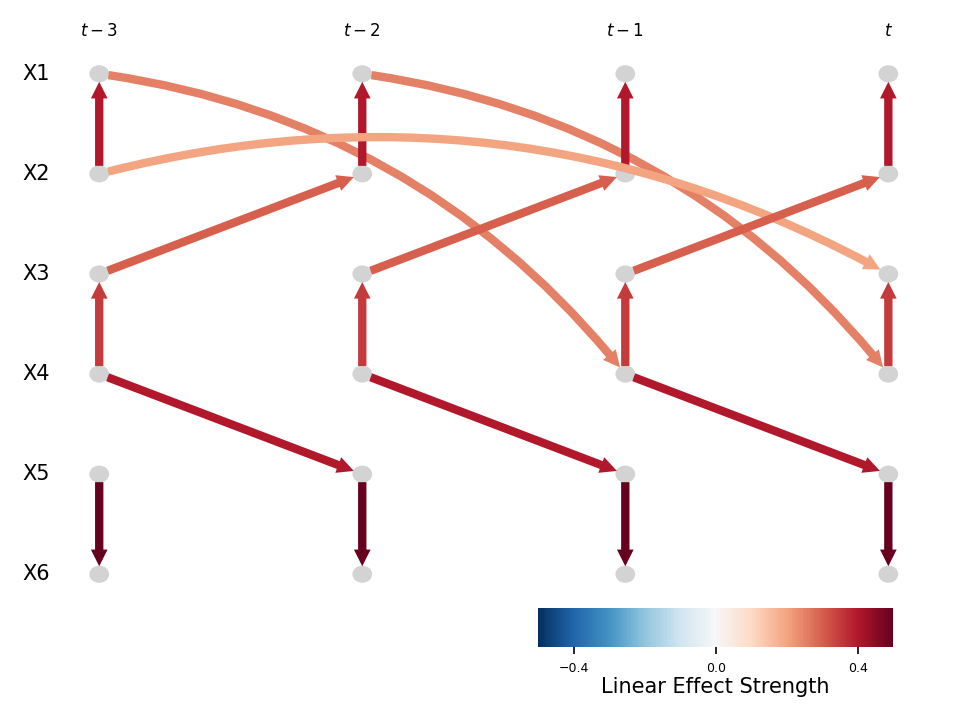}%
        \Description{Configuration 5: Causal graph with 6 variables, lag 3, 500 samples.}%
        \label{fig:cfg5}}
    \hfill
    \subfloat[Configuration 6: \(n=500\), vars=6, lag=4.]{%
        \includegraphics[width=0.3\linewidth]{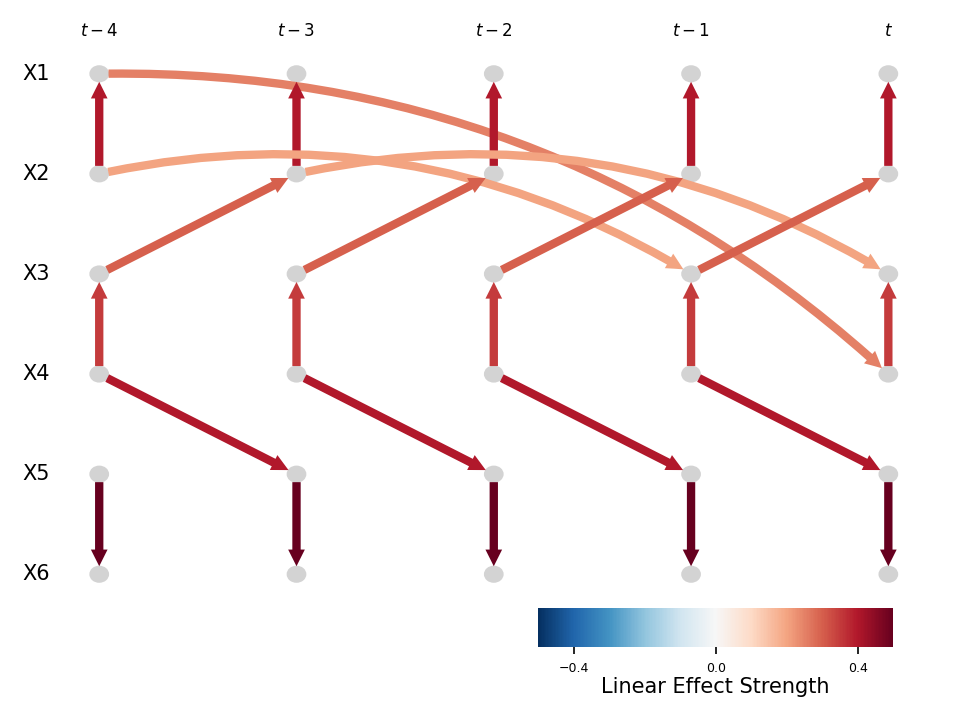}%
        \Description{Configuration 6: Causal graph with 6 variables, lag 4, 500 samples.}%
        \label{fig:cfg6}}

    \vspace{1em}

    % Third row
    \subfloat[Configuration 7: \(n=500\), vars=8, lag=2.]{%
        \includegraphics[width=0.3\linewidth]{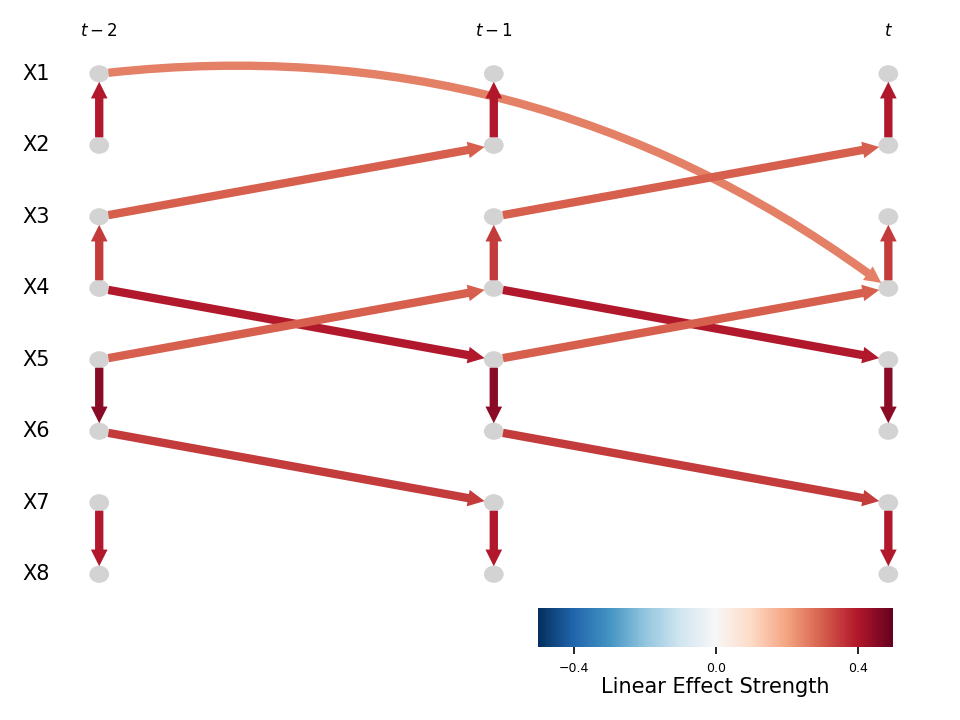}%
        \Description{Configuration 7: Causal graph with 8 variables, lag 2, 500 samples.}%
        \label{fig:cfg7}}
    \hfill
    \subfloat[Configuration 8: \(n=500\), vars=8, lag=3.]{%
        \includegraphics[width=0.3\linewidth]{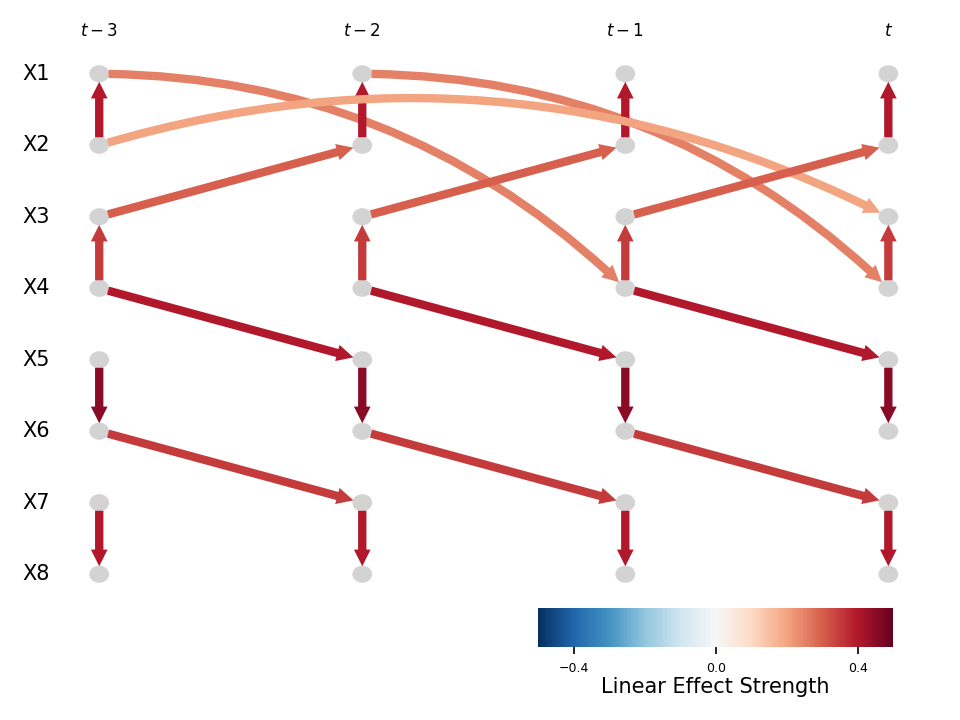}%
        \Description{Configuration 8: Causal graph with 8 variables, lag 3, 500 samples.}%
        \label{fig:cfg8}}
    \hfill
    \subfloat[Configuration 9: \(n=500\), vars=8, lag=4.]{%
        \includegraphics[width=0.3\linewidth]{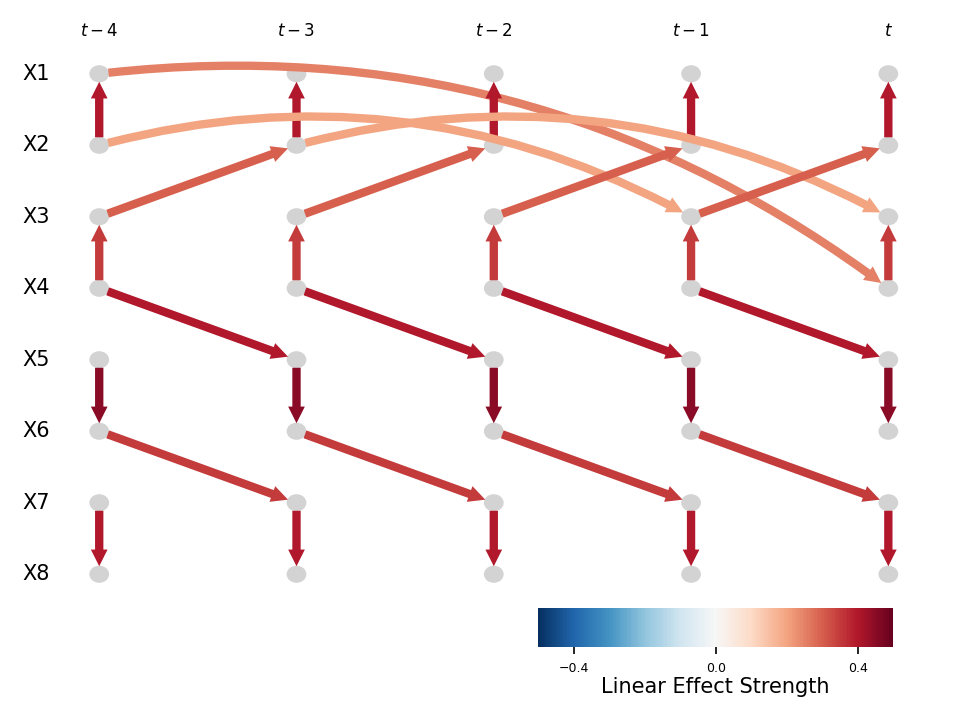}%
        \Description{Configuration 9: Causal graph with 8 variables, lag 4, 500 samples.}%
        \label{fig:cfg9}}

    \caption{Combined linear causal graphs for nine configurations (without confounders). Note that the underlying causal structure remains identical across different sample sizes (e.g., \(n=500\), \(1000\), \(3000\), and \(5000\)) and error distributions (Gaussian or Student-\(t\)) when no confounders are present. Only the statistical power varies with sample size and error distribution, while the graph topology remains unchanged.}
    \label{fig:combined-linear}
\end{figure}

\textbf{Nonlinear Time Series with Irregular Sampling and Mixed Errors (B2, B2C).}  
B2 datasets introduce irregularly sampled time series, where timestamps are generated using an exponential waiting time model, leading to non-uniform observation intervals. Many real-world datasets suffer from irregular sampling due to missing sensor readings, asynchronous logging, or event-based sampling \citep{raftery1992many}. This dataset is particularly useful for evaluating methods that assume regularly spaced time series data, such as vector autoregressive models, which may fail when faced with temporally misaligned data points.

Additionally, the error terms in B2 are sampled from a mixture of Gaussian and Laplace distributions, introducing heteroskedasticity and non-Gaussian noise characteristics. This ensures that methods relying on standard assumptions of normally distributed residuals will struggle, particularly those based on Gaussian Process modeling or least-squares regression \citep{yu2016temporal, xu2022inferring}. The confounded version (B2C) further compounds these challenges by introducing an unobserved confounder \( U \), which influences multiple observed variables nonlinearly. 

The presence of mixed noise makes it difficult to disentangle true causal relationships from artifacts introduced by variations in error distributions. Moreover, irregular sampling combined with mixed noise can lead to biased estimates in traditional causal discovery approaches, as standard estimators often assume homoscedastic and identically distributed errors \citep{schulam2017reliable}. Addressing these issues requires specialized methodologies that can simultaneously model temporal misalignment, mixed noise distributions, and confounding effects, making B2 and B2C valuable benchmarks for robust causal discovery methods.

\textbf{Nonlinear Time Series with Trigonometric Functions and Trend-Seasonal Components (C1, C1C).}  
The C1 datasets simulate time series with periodic behavior by incorporating trigonometric transformations alongside deterministic trends and seasonal components. Many real-world processes, including economic cycles \citep{hamilton2020time}, stock prices \citep{fama1970efficient}, and biological rhythms \citep{refinetti2016circadian}, exhibit strong periodic patterns. The dataset includes lagged dependencies modeled through sine and cosine transformations, with seasonal variations introduced via multiple harmonic functions \citep{chatfield2019analysis}. This variant is essential for testing methods designed for time series with seasonal adjustments, particularly in applications where trend-seasonality decomposition is crucial for identifying underlying causal influences \citep{cleveland1990stl}. The inclusion of deterministic trends ensures that the dataset also captures slow-moving structural changes, which are common in macroeconomic \citep{stock2016macroeconomic} and climate-related data \citep{mann1999global}.

The confounded version (C1C) introduces a latent variable \( U \), which interacts nonlinearly with selected variables, simulating real-world confounding where an unmeasured external factor drives observed seasonality. This can be seen in macroeconomic data, where external regulatory policies influence multiple economic indicators but are not explicitly recorded in datasets \citep{bernanke2005global}. Similarly, in climate science, long-term global trends such as atmospheric CO\(_2\) levels or oceanic temperature fluctuations may act as hidden confounders affecting seasonal weather patterns \citep{ipcc2021climate}. When such latent variables are not accounted for, standard seasonal adjustment techniques may produce misleading inferences, falsely attributing causal relationships to observed variables that are actually driven by unmeasured external forces \citep{shumway2017time}. The C1 and C1C datasets provide a rigorous benchmark for evaluating the robustness of causal discovery algorithms in detecting true seasonally dependent causal relationships while accounting for underlying confounding effects.

\textbf{Nonlinear Time Series with Irregular Sampling (C2, C2C).}  
C2 datasets extend C1 by introducing irregular sampling, where observation times are unevenly spaced due to an exponential waiting time model. This setup captures real-world scenarios where data points arrive at unpredictable intervals, such as in healthcare monitoring \citep{schulam2017reliable}, high-frequency financial data \citep{andersen2001distribution}, or event-driven systems \citep{reinhart2018review}. Many real-world sensor networks record measurements asynchronously, leading to irregular sampling that can disrupt time-lagged dependencies and obscure underlying causal relationships \citep{gama2013real}. When gaps between observations vary significantly, conventional time series models that assume uniform sampling may struggle to recover true causal effects, requiring adaptive methods that can account for missing or misaligned data points \citep{lipsitz1998using, little2019}.

The confounded variant (C2C) includes an unobserved variable \( U \), introducing additional dependencies that create spurious correlations, further complicating causal structure identification. This dataset is particularly relevant for applications such as medical diagnosis, where missing or irregularly recorded vitals can obscure causal relationships between physiological variables \citep{silva2012predicting}. For instance, in intensive care unit (ICU) monitoring, some vitals are recorded continuously, while others, such as blood tests, may only be available at irregular intervals, leading to potential biases when estimating causality \citep{ghassemi2015multivariate}. Additionally, irregular sampling combined with hidden confounders can introduce selection bias, where observed relationships differ significantly from the true underlying causal structure \citep{bareinboim2016causal}. The challenges posed by C2 and C2C emphasize the need for robust causal discovery techniques that can handle time-varying dependencies, irregular sampling gaps, and hidden confounding factors simultaneously.

\textbf{Linear Time Series with MCAR Missing Data (D1, D1C).}  
D1 datasets evaluate the impact of missing data on causal discovery methods by introducing \textit{Missing Completely At Random} (MCAR) mechanisms. The base dataset follows a linear autoregressive structure, with missing values introduced at predefined rates (ranging from 10\% to 30\%). This dataset is essential for studying the robustness of causal discovery algorithms when handling incomplete observations, a common challenge in financial modeling, epidemiology, and sensor networks \citep{little2019, rubin1976}. In many real-world settings, missing data occur due to device failures, inconsistent data collection protocols, or limitations in data storage capacity, leading to potential biases in causal estimates \citep{seaman2013handling}. By simulating MCAR conditions, D1 provides a controlled environment to assess how well causal discovery algorithms can recover the true underlying structure when data are missing in a completely random manner.

In the confounded version (D1C), an unobserved variable \( U \) is introduced, affecting multiple variables and making it more challenging to distinguish between true causal dependencies and correlations arising due to missing data patterns. The presence of latent confounding under MCAR missingness mimics real-world situations where key explanatory factors are either unmeasured or systematically absent from the dataset \citep{daniel2012causal}. For example, in clinical trials, patients might drop out randomly due to unrelated reasons, while an unmeasured genetic predisposition may simultaneously influence both treatment outcomes and dropout probabilities. Such scenarios can lead to misleading causal relationships if methods fail to adjust for the confounding effect \citep{bareinboim2016causal}. Addressing these challenges requires robust techniques capable of handling both missingness and unobserved confounding, making D1 and D1C valuable test cases for evaluating modern causal discovery methods.

\begin{table*}[ht!]
\centering
\caption{Causal discovery performance (TPR, FDR, SHD) for FGES, PC, PCMCI+, and LPCMCI on ten synthetic time-series datasets. 
Each dataset has 4 variables, a maximum lag of 2, and 500 samples. 
A dash (\texttt{--}) indicates no result reported.}
\label{tab:results}
\resizebox{\textwidth}{!}{%
\begin{tabular}{l|ccc|ccc|ccc|ccc}
\hline
\multirow{2}{*}{\textbf{Dataset}} & \multicolumn{3}{c|}{\textbf{FGES}} & \multicolumn{3}{c|}{\textbf{PC}} & \multicolumn{3}{c|}{\textbf{PCMCI+}} & \multicolumn{3}{c}{\textbf{LPCMCI}} \\
 & \textbf{TPR} & \textbf{FDR} & \textbf{SHD} & \textbf{TPR} & \textbf{FDR} & \textbf{SHD} & \textbf{TPR} & \textbf{FDR} & \textbf{SHD} & \textbf{TPR} & \textbf{FDR} & \textbf{SHD} \\
\hline
\texttt{A1 (Linear with Gaussian error)} 
    & 0.56 & 0.29 & 6.00
    & \textbf{0.67} & 0.25 & \textbf{5.00}
    & \textbf{1.00} & \textbf{0.00} & \textbf{0.00}
    & 0.78 & 0.13 & \textbf{3.00} \\

\texttt{A1 (Linear with Student's t error)} 
    & 0.67 & 0.25 & 5.00
    & \textbf{0.67} & 0.33 & 6.00
    & 0.67 & 0.33 & 6.00
    & 0.67 & 0.14 & 4.00 \\

\texttt{B1 (Nonlinear (polynomial) with Gaussian error)} 
    & 0.00 & \textbf{0.00} & 9.00
    & 0.00 & \textbf{0.00} & 9.00
    & 0.00 & 1.00 & 10.00
    & 0.00 & 1.00 & 12.00 \\

\texttt{B1 (Nonlinear (polynomial) with Student's t error)} 
    & 0.33 & 1.00 & 29.00
    & 0.44 & 0.67 & 13.00
    & 0.33 & 0.93 & 45.00
    & --   & --   & --   \\

\texttt{C1 (Nonlinear with trend and seasonality and Gaussian error)} 
    & 0.44 & 0.84 & 26.00
    & 0.44 & 0.78 & 19.00
    & 0.00 & 1.00 & 26.00
    & 0.00 & 1.00 & 32.00 \\

\texttt{A1C (A1 with unobserved confounder and Gaussian error)} 
    & \textbf{0.78} & 0.30 & 5.00
    & 0.56 & 0.38 & 7.00
    & 0.67 & 0.50 & 9.00
    & \textbf{1.00} & 0.25 & \textbf{3.00} \\

\texttt{A1C (A1 with unobserved confounder and Student's t error)} 
    & 0.67 & \textbf{0.00} & \textbf{3.00}
    & 0.44 & 0.43 & 8.00
    & 0.67 & 0.50 & 9.00
    & 0.67 & 0.33 & 6.00 \\

\texttt{B1C (B1 with unobserved confounder and Gaussian error)} 
    & 0.00 & \textbf{0.00} & 9.00
    & 0.00 & \textbf{0.00} & 9.00
    & 0.22 & \textbf{0.00} & 7.00
    & 0.22 & \textbf{0.00} & 7.00 \\

\texttt{B1C (B1 with unobserved confounder and Student's t error)} 
    & 0.00 & \textbf{0.00} & 9.00
    & 0.00 & \textbf{0.00} & 9.00
    & 0.33 & 0.94 & 51.00
    & 0.11 & 0.97 & 36.00 \\

\texttt{C1C (C1 with unobserved confounder and Gaussian error)} 
    & 0.44 & 0.84 & 26.00
    & 0.33 & 1.00 & 20.00
    & 0.33 & 0.79 & 17.00
    & 0.11 & 0.93 & 22.00 \\
\hline
\end{tabular}
}
\end{table*}

\textbf{Nonlinear Time Series with Block Missingness (D2, D2C).}  
D2 datasets extend nonlinear time series by introducing structured block missingness, where missing values occur in contiguous segments rather than randomly. This setup simulates real-world data collection issues, such as sensor failures, dropout in longitudinal studies, or systematic gaps in survey responses \citep{little2019}. Unlike standard missing data mechanisms that assume independent missingness across observations, block missingness presents additional challenges by introducing dependencies in missing patterns over time \citep{robins2000analysis}. 

Missing blocks are introduced probabilistically, following both the Missing at Random (MAR) and Not Missing at Random (NMAR) assumptions \citep{rubin1976}. Under MAR, the probability of missingness depends only on observed values, whereas under NMAR, missingness depends on unobserved variables, making standard imputation and estimation techniques potentially biased \citep{daniel2012causal}. This distinction is crucial, as many real-world datasets exhibit NMAR behavior, particularly in healthcare and economic time series where systematic missingness is linked to external factors like policy changes or health deterioration.

The confounded version (D2C) further complicates inference by adding an unobserved confounder \( U \), which affects multiple variables and creates an intricate missingness-causality interplay. The presence of \( U \) introduces additional dependencies between missing values and causal structures, leading to spurious correlations and biased causal estimates if not properly accounted for \citep{bareinboim2016causal}. These datasets serve as a benchmark for evaluating causal discovery methods' ability to handle structured missingness while mitigating the effects of latent confounding.

\textbf{Nonlinear Time Series with Mixed Errors and Mixed Missing Patterns (D3, D3C).}  
The D3 datasets introduce a combination of challenging conditions, including mixed noise distributions, irregular sampling, and hybrid missing data mechanisms. The noise model blends Gaussian and Laplace distributions with a tunable mixing ratio, allowing for controlled variations in heavy-tailed errors \citep{kotz2012laplace}. Observation timestamps follow a non-uniform distribution, making time-lagged relationships difficult to reconstruct \citep{reinhart2018review, gama2013real}. 

The confounded version (D3C) extends this by incorporating an unobserved confounder \( U \), which exerts nonlinear influences on selected variables. This dataset variant is particularly useful for studying the combined effects of irregular sampling, missing data, and unobserved confounding, providing a rigorous benchmark for real-world causal discovery \citep{bareinboim2016causal, robins2000analysis}. The simultaneous presence of mixed noise and structured missingness significantly increases the difficulty of distinguishing genuine causal effects from statistical artifacts. Furthermore, the interplay between missing patterns and latent confounding may lead to spurious causal links, requiring advanced methodologies capable of handling both incomplete and biased observational data \citep{daniel2012causal, little2019}.

\textbf{Implementation and Evaluation.}  
All datasets are implemented with automated causal structure extraction, supporting visualization of causal graphs, time series trajectories, and missing data patterns using the \texttt{Tigramite} Python library. By providing a diverse range of datasets that capture real-world challenges, TimeGraph serves as a rigorous benchmark for evaluating causal discovery algorithms under varying conditions of complexity, irregularity, and confounding.

\section{Realism and Benchmarking Utility of the Synthetic Datasets} \label{sec:realism}

Our synthetic datasets are designed to capture many of the complexities observed in real-world time series while retaining a known ground-truth causal structure. This dual characteristic enables a rigorous and objective evaluation of causal discovery algorithms. In particular, our datasets mirror several critical properties observed in empirical data:

\begin{itemize}
    \item \textbf{Temporal Dynamics:} Real-world time series often exhibit nonstationarity due to trends and seasonal effects, as well as irregular sampling intervals. Our datasets incorporate deterministic trends, seasonal oscillations (modeled via sine and cosine functions), and non-uniform time grids generated from exponential distributions \cite{granger1969investigating, runge2019}. Figure~\ref{fig:c1_and_stock} shows a side-by-side comparison between our synthetic dataset (C1 variant) and real-world stock data. The synthetic time series captures an overall upward drift, periodic fluctuations, and variability in amplitude, much like the real stock market data. For a detailed configuration of the C1 dataset, see Table~\ref{tab:synthetic-datasets}.
    
    \item \textbf{Noise Characteristics:} To reflect heterogeneous error structures, we simulate data using multiple noise models. Some variants use additive Gaussian noise (\(\mathcal{N}(0, 0.1^2)\)), while others use heavy-tailed noise (Student’s t-distribution with 3 degrees of freedom) or a mixed noise model that probabilistically combines Gaussian and Laplace noise. This approach is motivated by findings that real-world errors are often non-Gaussian \citep{mooij2016distinguishing, zhang2012kernel}.

\begin{figure*}[ht]
    \centering
    \includegraphics[width=\linewidth]{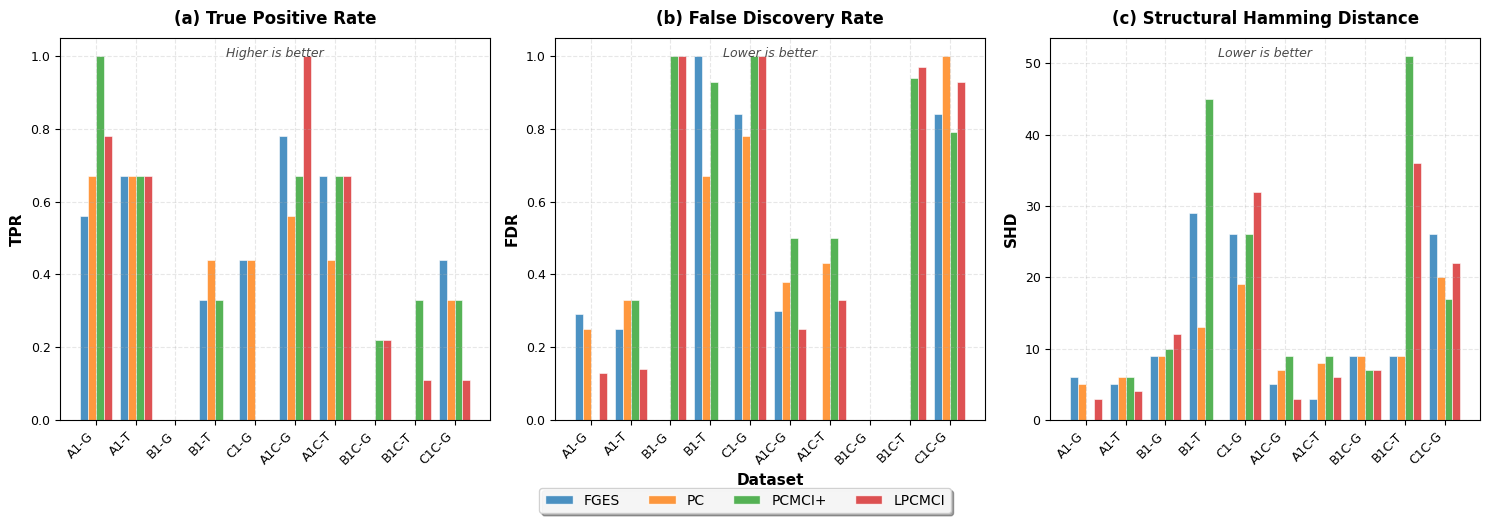}
    \Description{Bar charts comparing the performance of FGES, PC, PCMCI+, and LPCMCI algorithms on synthetic time-series datasets. Metrics include True Positive Rate, False Discovery Rate, and Structural Hamming Distance, showing results for both linear and nonlinear settings, with and without confounders.}
    \caption{Comparative bar charts illustrating the performance of four causal discovery algorithms (FGES, PC, PCMCI+, LPCMCI) across various synthetic time-series datasets. Each subplot shows a different metric (True Positive Rate, False Discovery Rate, and Structural Hamming Distance), allowing a side-by-side evaluation of how each method performs in the presence of linear vs.\ nonlinear relations and potential unobserved confounders.}
    \label{fig:comparison}
\end{figure*}

    \item \textbf{Missing Data Patterns:} Recognizing that real-world data often suffer from both random sensor dropouts (MCAR) and systematic block missingness (MAR/NMAR), we simulate these conditions explicitly. Our implementation randomly drops data points or removes contiguous blocks of data, following principles discussed in the missing data literature \cite{little2019}. This approach challenges causal discovery methods to recover underlying structures despite incomplete observations.
    
    \item \textbf{Latent Confounding:} Unobserved confounders can obscure true causal relationships. Several of our dataset variants include an unobserved variable \(U\) that is integrated into the structural equations (typically via quadratic terms) to model hidden influences, a common challenge in empirical causal structure identification \cite{pearl2009causality}.
\end{itemize}

Because the ground-truth causal graph is fully known in our synthetic setup, these datasets provide an ideal environment for evaluating causal discovery algorithms under realistic yet controlled conditions. Researchers can assess algorithmic performance using standard metrics (e.g., True Positive Rate, False Discovery Rate, Structural Hamming Distance) and systematically explore how variations in temporal dynamics, noise, missing data, and confounding affect the recovery of causal structure. This benchmarking framework thereby bridges the gap between idealized synthetic benchmarks and the multifaceted nature of real-world data, supporting more robust, generalizable, and reproducible advancements in causal discovery research.

\textbf{The Trade-Off: Realism vs. Control.} It is important to acknowledge the inherent trade-off in any synthetic benchmark: capturing the full spectrum of real-world complexity versus maintaining a controlled environment with known ground truth. While TimeGraph aims to mirror many realistic properties, its primary objective is to provide a setting where causal structures are \textit{fully known}, allowing for objective, metrics-based evaluation. This controlled approach enables researchers to precisely dissect \textit{why} algorithms succeed or fail under specific conditions—like trends, non-stationarity, or confounding —an analysis often impossible with empirical data. Consequently, while we strive for plausible dynamics (as shown in Figure~1), perfect distributional replication is secondary to the goal of providing a \textit{robust and verifiable testbed for causal structure learning}.

\section{Benchmark and Results} \label{sec:benchmark}

We evaluate four causal discovery algorithms ( FGES~\citep{chickering2002optimal, ramsey2017million}, PC~\citep{spirtes2000causation}, PCMCI+~\citep{runge2020discovering}, and LPCMCI~\citep{gerhardus2020high}) on ten synthetic time-series datasets, each containing 4 observed variables (\(\text{vars}=4\)), a maximum lag of 2 (\(\text{lag}=2\)), and 500 samples. We track three standard metrics in causal discovery:

\begin{itemize}
    \item \textbf{TPR (True Positive Rate):} Fraction of ground-truth edges that are discovered.
    \item \textbf{FDR (False Discovery Rate):} Fraction of discovered edges that are not present in the ground truth.
    \item \textbf{SHD (Structural Hamming Distance):} The total number of edge additions, deletions, or direction reversals required to transform the learned graph into the true structure.
\end{itemize}

\subsection{Baseline Approaches}

\paragraph{FGES}
Fast Greedy Equivalence Search (FGES)~\citep{chickering2002optimal, ramsey2017million} greedily searches over equivalence classes of directed acyclic graphs (DAGs). It uses a score-based approach (often with linear-Gaussian assumptions) to identify candidate edges. FGES can work well when data are sufficiently large and approximately linear, but strong nonlinearity or heavy-tailed noise may undermine its performance.

\paragraph{PC}
The PC algorithm~\citep{spirtes2000causation} is a constraint-based method that systematically removes edges from a fully connected graph using conditional independence tests (often partial correlation). It assumes faithfulness and tends to be computationally efficient for moderate graph sizes. However, PC may fail to capture nonlinear or non-Gaussian effects if only linear independence tests are used.

\paragraph{PCMCI+}
PCMCI+~\citep{runge2020discovering} extends the PC approach to time-series data by incorporating lagged variables and the notion of momentary conditional independence (MCI). It iteratively refines candidate parents and leverages robust independence tests. While PCMCI+ can capture certain autocorrelated structures, it may still produce spurious edges under strong nonlinearities or confounded scenarios if the test assumptions are not satisfied.

\paragraph{LPCMCI}
LPCMCI~\citep{gerhardus2020high} adds further adaptations for partially unobserved confounders, aiming to reduce biases from hidden variables. Its layered structure of independence tests attempts to filter out false positives introduced by latent factors. Although LPCMCI can handle some complex dependencies, strong nonlinearity remains an open challenge.

\subsection{Results}

It is important to interpret these results within their intended context. This evaluation aims primarily to demonstrate TimeGraph's utility in highlighting performance variations among established algorithms when faced with specific challenges like nonlinearity and confounding. It is not intended as an exhaustive, large-scale comparison but rather as a proof-of-concept for the benchmark's effectiveness.

Figure~\ref{fig:comparison} and Table~\ref{tab:results} summarize the True Positive Rate (TPR), False Discovery Rate (FDR), and Structural Hamming Distance (SHD) for each causal discovery method across different dataset variants. We evaluate linear cases (\texttt{A1.*})—both unconfounded and confounded (\texttt{A1C.*})—alongside nonlinear datasets (\texttt{B1.*}, \texttt{C1.*}) that incorporate polynomial transformations or \textit{t}-distributed noise, often with confounding (\texttt{B1C.*}, \texttt{C1C.*}). 

Results show that linear Gaussian datasets yield higher TPR and lower FDR/SHD across FGES, PC, PCMCI+, and LPCMCI. These methods perform well in structured settings where relationships adhere to linear assumptions, allowing them to accurately recover causal edges while minimizing false discoveries. However, in purely nonlinear datasets, TPR often drops to 0, while FDR approaches 1.0, indicating that none of the tested methods consistently capture strong nonlinear interactions.

As illustrated in Figure~\ref{fig:comparison}, confounding generally degrades performance, especially in combination with nonlinearity (\texttt{B1C.*}, \texttt{C1C.*}). While LPCMCI is designed to mitigate latent confounding effects, its TPR remains low in heavily nonlinear settings, and FDR can spike significantly. In simpler linear scenarios, LPCMCI performs well by suppressing spurious edges (e.g., TPR~\(=1.0\), FDR~\(=0.25\) on \texttt{A1C} with Gaussian errors), demonstrating its potential when model assumptions align closely with the underlying data structure.

Linear or near-linear datasets (\texttt{A1} with Gaussian and \textit{t}-distributed errors) favor FGES and PC, which assume partial correlations and parametric simplicity. PCMCI+ and LPCMCI also perform well in these cases, often recovering nearly all true causal edges. However, in purely nonlinear settings (e.g., \texttt{B1} with Gaussian errors, \texttt{C1} with Gaussian errors), TPR frequently drops to zero, and FDR reaches near-random levels. Even LPCMCI, which is explicitly designed to handle confounding, exhibits only partial success in certain cases (\texttt{A1C} with Gaussian errors) and struggles when strong nonlinearities are present (\texttt{C1C} with Gaussian errors).

These results suggest that the effectiveness of causal discovery methods is highly dependent on the structural properties of the data. In settings with strong nonlinearity or significant confounding, standard methods often fail, underscoring the need for more specialized or robust approaches capable of adapting to complex causal dependencies. This highlights the critical role of comprehensive benchmarks like TimeGraph, which allow for the systematic identification of these failure points under controlled conditions, ensuring fair and transparent assessment. By enabling researchers to test algorithms across such diverse scenarios, our work aims to provide a robust foundation for developing next-generation methods truly capable of handling real-world complexities.

\section{Guidelines for Use and Best Practices} \label{sec:guidelines}

It is recommended that dataset variants from TimeGraph be selected based on the specific research objectives. For example, variants such as A2, B2, and C2 are suitable for investigations into nonstationarity and irregular sampling, while paired comparisons (for instance, A1 versus A1C) facilitate the evaluation of latent confounding effects (see Table~\ref{tab:synthetic-datasets}). All experimental settings including maximum lag, noise parameters, sampling schemes, and missing data patterns should be documented using the provided data-generation scripts.

Benchmarking should employ standard evaluation metrics like TPR, FDR, and SHD, with results reported on both complete and observed data when missingness occurs. Full transparency and reproducibility are essential; thus, it's strongly encouraged to share all code, configuration files, and detailed protocols for independent verification. We must acknowledge that although TimeGraph captures many real-world properties, synthetic data inevitably have limitations compared to empirical datasets \cite{runge2019}.

Finally, community contributions are welcomed to extend TimeGraph by proposing new datasets, improving data-generation scripts, and suggesting additional evaluation metrics \cite{mooij2016distinguishing}. Such collaborative efforts will enhance the robustness and generalizability of causal discovery methods and promote reproducible research in this field.

\section{Limitations} \label{sec:limitations}

Although the TimeGraph suite incorporates many realistic characteristics, we acknowledge several limitations beyond its synthetic nature. Firstly, the empirical evaluations focus on a representative but limited set of algorithms and operate at a moderate scale (e.g., up to 8 variables), which may not fully reflect challenges in high-dimensional settings. Secondly, TimeGraph exclusively generates observational data. While fundamental, the absence of interventional data limits its use for benchmarking algorithms designed for active learning or estimating causal effects under interventions. Thirdly, while the functional forms and missing data mechanisms are diverse, they cannot capture every real-world nuance \cite{runge2019, little2019}. For example, abrupt regime shifts or event-driven dynamics are not explicitly modeled, and latent confounders, though implemented, simplify the complexity of hidden factors in real applications. Finally, the current public release consists of generation scripts; a full, user-friendly Python library with enhanced flexibility is planned as future work.

\section{Conclusion and Future Directions} \label{sec:conclusion}

The pursuit of robust causal discovery methods for time-series data has long been hindered by the scarcity of suitable benchmarks that reflect real-world complexities while providing known ground truths. Addressing this gap, we introduced TimeGraph, a synthetic benchmark suite for time-series causal discovery that emulates key temporal characteristics such as trends, seasonality, irregular sampling, diverse noise patterns, and structured missingness. TimeGraph supports both linear and nonlinear models, including polynomial and trigonometric transformations, and offers variants with and without latent confounders. Empirical evaluations show that algorithm performance is highly sensitive to noise type, sampling irregularity, and missing data patterns.

Looking ahead, efforts are underway to extend TimeGraph and enhance its utility. Key priorities include: (1) Incorporating more complex dynamic models, such as regime shifts and time-varying causal structures; (2) Enhancing missing data mechanisms and developing modules for generating interventional data; (3) Expanding to larger-scale datasets and evaluating a broader range of algorithms, including deep learning methods; and (4) Releasing TimeGraph as a standalone Python library with flexible options for graph structure, parameter generation, sample size, and lag configuration. We believe these enhancements, supported by community contributions, will further advance research in time-series causal discovery.

\section*{Acknowledgments}
This work is supported by iHARP: NSF HDR Institute for Harnessing Data and Model Revolution in the Polar Regions (Award\# 2118285). The views expressed in this work do not necessarily reflect the policies of the NSF, and endorsement by the Federal Government should not be inferred.

\bibliographystyle{ACM-Reference-Format}
\bibliography{References}
\appendix

\end{document}